\documentclass{article}

\PassOptionsToPackage{square,sort,comma,numbers}{natbib}
 \usepackage[final]{nips_2017}

\usepackage[utf8]{inputenc} 
\usepackage[T1]{fontenc}    
\usepackage{hyperref}       
\usepackage{url}            
\usepackage{booktabs}       
\usepackage{amsfonts}       
\usepackage{nicefrac}       
\usepackage{microtype}      
\usepackage{amsmath}    
\usepackage{MnSymbol}
\usepackage{graphicx}
\usepackage{subcaption}
\usepackage{makecell}
\usepackage[most]{tcolorbox}
\usepackage{multirow}

\newcommand{\ud}{\mathrm{d}}
\newcommand{\R}{\mathbb{R}}

\newcommand{\I}{\mathbb{I}}

\title{Recurrent Estimation of Distributions}

\author{
  Junier B. Oliva\thanks{Equal Contribution.},\ \ Avinava Dubey$^*$\hspace{-1.7mm},\ \ Barnab\'{a}s P\'{o}czos,\ \ Eric P. Xing,\ \ Jeff Schneider \\
  Machine Learning Department\\
  Carnegie Mellon University\\
  Pittsburgh, PA 15213 \\
  \texttt{\{joliva, akdubey, bapoczos, epxing, schneide\}@cs.cmu.edu} \\
}

\begin{document}

\maketitle

\begin{abstract}
This paper presents the recurrent estimation of distributions (RED) for modeling real-valued data in a semiparametric fashion. RED models make two novel uses of recurrent neural networks (RNNs) for density estimation of general real-valued data. First, RNNs are used to transform input covariates into a latent space to better capture conditional dependencies in inputs. After, an RNN is used to compute the conditional distributions of the latent covariates. The resulting model is efficient to train, compute, and sample from, whilst producing normalized pdfs. 
The effectiveness of RED is shown via several real-world data experiments.  
Our results show that RED models achieve a lower held-out negative log-likelihood than other neural network approaches across multiple dataset sizes and dimensionalities.
Further context of the efficacy of RED is provided by considering  anomaly detection tasks, where we also observe better performance over alternative models.
\end{abstract}

\section{Introduction}
Density estimation is at the core of a multitude of machine learning applications. However, this fundamental task, which encapsulates the understanding of data, is difficult in the general setting due to issues like the curse of dimensionality. Furthermore, general data, unlike spatial/temporal data, does not contain apriori known correlations among covariates that may be exploited and engineered with. For example, image data has known correlations among neighboring pixels that may be hard-coded into a model, whereas one must find such correlations in a data-driven fashion with general data. 

\begin{figure}
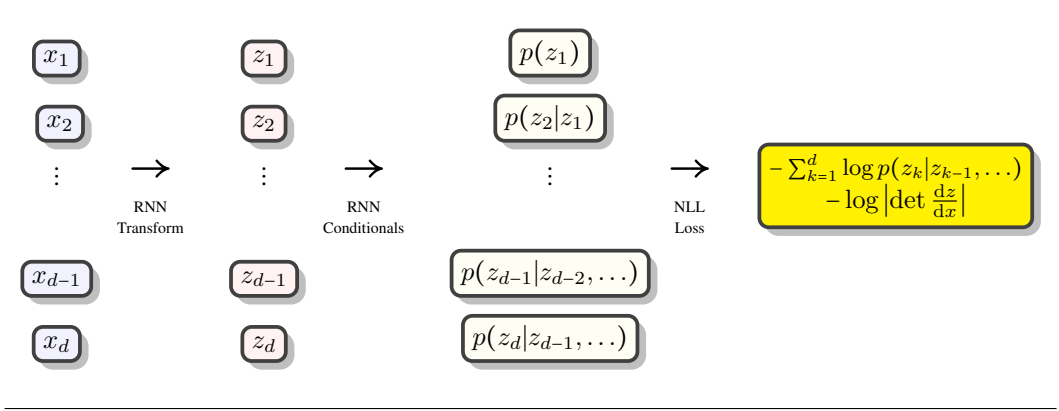

\begin{center}
\begin{tabular}{ c c c c c c l } 
  \hline\\ 
  \tcbox[top=0pt,left=0pt,right=0pt,bottom=0pt,enhanced,drop shadow,colback=blue!5!]{$x_1$} &  & \tcbox[top=0pt,left=0pt,right=0pt,bottom=0pt,enhanced,drop shadow,colback=red!5!]{$z_{1}$} & & \tcbox[top=0pt,left=0pt,right=0pt,bottom=0pt,enhanced,drop shadow,colback=yellow!5!]{$p(z_1)$} & &
  \vspace{1mm}\\ 
  \vspace{-5mm}
  \tcbox[top=0pt,left=0pt,right=0pt,bottom=0pt,enhanced,drop shadow,colback=blue!5!]{$x_2$} & & \tcbox[top=0pt,left=0pt,right=0pt,bottom=0pt,enhanced,drop shadow,colback=red!5!]{$z_{2}$} & & \tcbox[top=0pt,left=0pt,right=0pt,bottom=0pt,enhanced,drop shadow,colback=yellow!5!]{$p(z_2 | z_1)$} & &
  \\ 
  $\vdots$ & 
  \hspace{-1mm}{\huge$\underset{
      \underset{\text{\tiny{Transform}}}{\text{\tiny{RNN}}}
    }
    {\rightarrow}$
  } & 
  $\vdots$ & 
  \hspace{-1mm}{\huge$\underset{
      \underset{\text{\tiny{Conditionals}}}{\text{\tiny{RNN}}}
    }
    {\rightarrow}$
  } & 
  $\vdots$ & 
  \hspace{-2mm}{\huge$\underset{
      \underset{\text{\tiny{Loss}}}{\text{\tiny{NLL}}}
    }
    {\rightarrow}$
  } &
  \makecell{\\\tcbox[top=0pt,left=0pt,right=0pt,bottom=0pt,enhanced,drop shadow,colback=yellow]{\makecell{\small$-\sum_{k=1}^d\log p(z_{k} | z_{k-1}, \ldots)$\\
    $-\log\left|\det\frac{\ud z}{\ud x}\right|$
  }}} \vspace{1mm}\\
  \tcbox[top=0pt,left=0pt,right=0pt,bottom=0pt,enhanced,drop shadow,colback=blue!5!]{$x_{d-1}$} &  & \tcbox[top=0pt,left=0pt,right=0pt,bottom=0pt,enhanced,drop shadow,colback=red!5!]{$z_{d-1}$} & & \tcbox[top=0pt,left=0pt,right=0pt,bottom=0pt,enhanced,drop shadow,colback=yellow!5!]{$p(z_{d-1} | z_{d-2}, \ldots)$} & &
  \vspace{1mm} \\ 
  \tcbox[top=0pt,left=0pt,right=0pt,bottom=0pt,enhanced,drop shadow,colback=blue!5!]{$x_{d}$} &  & \tcbox[top=0pt,left=0pt,right=0pt,bottom=0pt,enhanced,drop shadow,colback=red!5!]{$z_{d}$} & & \tcbox[top=0pt,left=0pt,right=0pt,bottom=0pt,enhanced,drop shadow,colback=yellow!5!]{$p(z_d | z_{d-1}, \ldots)$} & &
  \\ 
  \vspace{0.1cm}\\ \hline
\end{tabular}
\end{center}
  \caption{Illustration of the RED model. First, a transformation  of inputs $x$ is computed using RNNs to produce $z$. The transformed variable $z$ is then fed through an RNN to compute chain-rule conditionals $p(z_k | z_{k-1}, \ldots, z_1)$. Finally, the conditionals on $z$ are renormalized by a determinant of the Jacobian to yeild the negative log likelihood $-\log p(x) = -\log\left|\det\frac{\ud z}{\ud x}\right| -\sum_{k=1}^d\log p(z_{k} | z_{k-1}, \ldots)$. } \label{fig:model}
\end{figure}

In this paper we propose the method of recurrent estimation of distributions (RED) for general real-valued data. Our method will make use of recurrent neural networks (RNNs) to produce normalized pdf estimates that are efficient to sample from. We will show with the use of RNNs that we can produce transformations to exploit latent correlations in a data-driven fashion. At a high level our method consists of two applications of RNNs: (a) to perform a change of variables operation and (b) to produce parameters of conditional densities (see Figure~\ref{fig:model}). 

The following are our contributions. To the best of our knowledge, this is the first application of RNNs for estimating conditionals in general non-spatial/temporal data. Furthermore, this paper considers the first application of RNNs to perform change of variables for density estimation. These two ideas have been under-explored in previous general density estimation approaches despite their natural fit. Lastly, we make use of these novel concepts to develop a simple, efficient architecture for density estimation. 
%


The remainder of the paper is structured as follows. First, we describe our model in detail. After, we discuss related work and contrast our approach to previous methods. We then illustrate the efficacy of RED models with respect to 
other neural network
methods using test-likelihoods on several datasets. We also show that our model outperforms other density estimators on the task of anomaly detection on several benchmark datasets.

\section{Model}
Below we describe the RED model and provide intuition for its design. 
The RED model makes two crucial uses of recurrent neural networks: 
to produce output features that will be fed forward to yield the parameters of conditional densities,
and
to transform the dimensions of $x$ to a more adept space for density estimation.

\subsection{Recurrent Conditional Distributions}
First we discuss the use of an RNN for the estimation of conditional distributions. As motivation, consider the following analogy between sequence estimation and general density estimation. 

Suppose one has a sequence of multivariate points $y_1 \in \R^d, \ldots, y_T \in \R^d$;
a common task is to predict the distribution of the next point given the sequence seen thus far: $p(y_{t+1} | y_t, \ldots, y_1)$.
A nonparametric treatment of this estimation task will encounter the curse of dimensionality due to the ever-growing dimensionality of ``inputs'' $y_t, \ldots , y_1$.
Furthermore, a completely general approach will require a multitude of different estimation tasks, one for each set of past points: $p(y_{t+1} | y_t, \ldots, y_1)$, $p(y_{t} | y_{t-1}, \ldots, y_1)$, $\ldots$, $p(y_{2} | y_1)$, $p(y_1)$. 
RNNs address both the issue of the curse of dimensionality in the inputs and the issue of multiple prediction tasks through hidden states.

That is, RNNs model conditional distributions as a function of the past points: $p(y_{t+1} | y_t, \ldots, y_1) = p(y_{t+1} | f(y_t, \ldots, y_1)) = p(y_{t+1} | h_t)$ where $h_t$ is the RNN state after seeing the past $t$ points. By modeling the conditional distribution as a parametric function of past points, RNNs are able to avoid the curse of dimensionality present in nonparametric estimators. Furthermore, with its dependency on prior states, $h_t = g(y_t, h_{t-1})$, an RNN approach provides a ``weight-sharing'' scheme across the various conditionals $p(y_{t+1} | h_t)$, $p(y_{t} | h_{t-1})$, $\ldots$. This approach is akin to common multi-task learning techniques, and alleviates the problem of solving for many separate tasks in a general sequential estimation method (without a Markovian assumption). 

\begin{figure}[h]
\begin{center}
\vspace{0.25cm}
\begin{tabular}{ c c c c c } 
  \hline\\ 
  \tcbox[top=0pt,left=0pt,right=0pt,bottom=0pt,enhanced, drop shadow,colback=white]
  {$\emptyset$} & \makecell{\huge$\rightarrow$} & 
  \tcbox[top=0pt,left=0pt,right=0pt,bottom=0pt,enhanced,sharp corners, drop shadow,colback=red!5!]{$h_0$} &  {\huge$\rightarrow$} & 
  \tcbox[top=0pt,left=0pt,right=0pt,bottom=0pt,enhanced,drop shadow,colback=yellow!1!]{
    $p(x_1) = p(x_1 | h_{0})$ 
    \makecell{\includegraphics[width=0.1\textwidth]{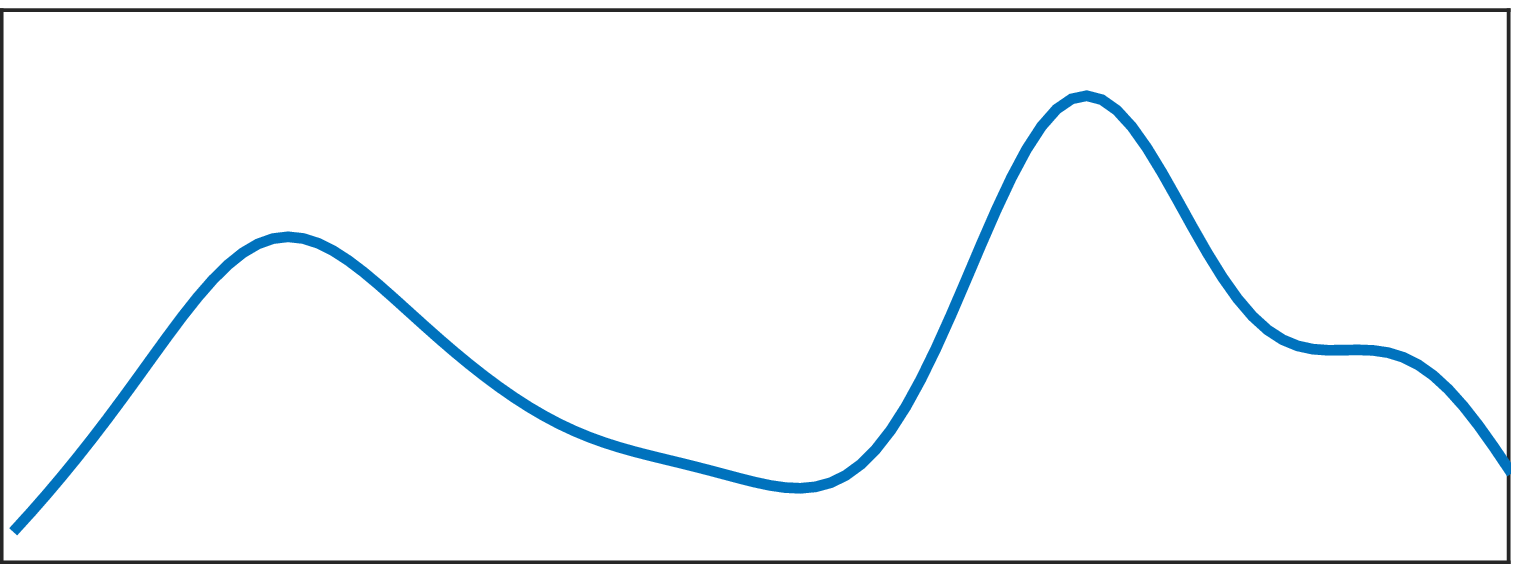}}
    }
  \\ 
  \vspace{-2mm} & & {\LARGE$\downarrow$} & &  \\ 
  \tcbox[top=0pt,left=0pt,right=0pt,bottom=0pt,enhanced,drop shadow,colback=blue!5!]{$x_1$} & {\huge$\rightarrow$} & \tcbox[top=0pt,left=0pt,right=0pt,bottom=0pt,enhanced,sharp corners, drop shadow,colback=red!5!]{$h_1$} &  {\huge$\rightarrow$} & 
  \tcbox[top=0pt,left=0pt,right=0pt,bottom=0pt,enhanced,drop shadow,colback=yellow!1!]{
    $p(x_2 | x_1)=p(x_2 | h_{1})$
    \makecell{\includegraphics[width=0.1\textwidth]{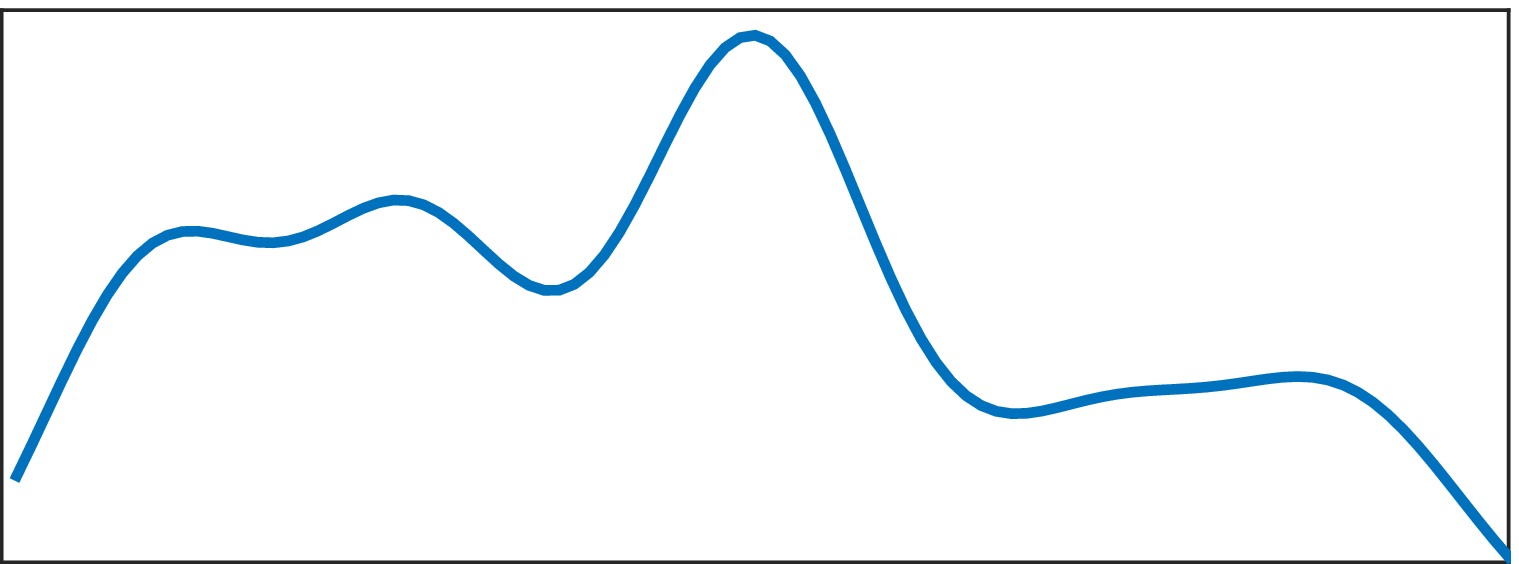}}
  }
  \\ 
  \vspace{-2mm} & & {\LARGE$\downarrow$} & &  \\
  \tcbox[top=0pt,left=0pt,right=0pt,bottom=0pt,enhanced,drop shadow,colback=blue!5!]{$x_2$} & {\huge$\rightarrow$} & \tcbox[top=0pt,left=0pt,right=0pt,bottom=0pt,enhanced,sharp corners, drop shadow,colback=red!5!]{$h_2$} &  {\huge$\rightarrow$} & 
  \tcbox[top=0pt,left=0pt,right=0pt,bottom=0pt,enhanced,drop shadow,colback=yellow!1!]{
    $p(x_3 | x_2, x_1)=p(x_3 | h_{2})$ 
    \makecell{\includegraphics[width=0.1\textwidth]{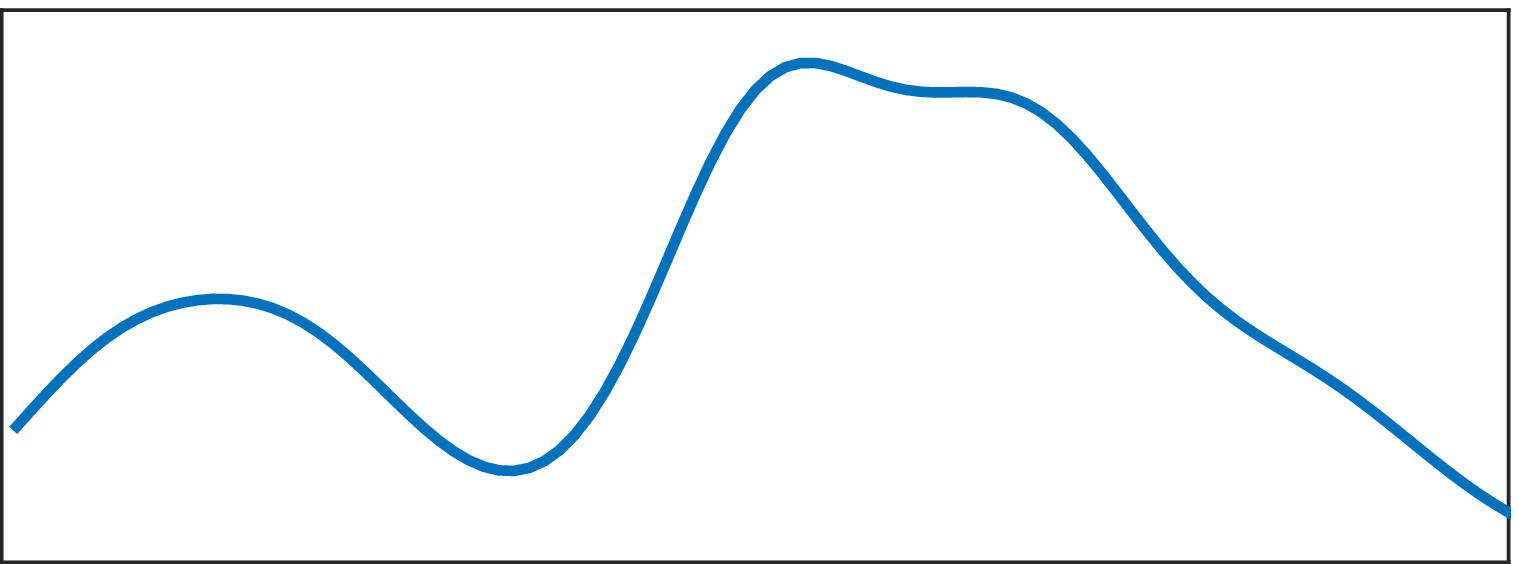}}
  }
  \\ 
   & & {\LARGE$\downarrow$} & &  \\
  \vspace{-2mm} & & $\vdots$ & &  \\
  \tcbox[top=0pt,left=0pt,right=0pt,bottom=0pt,enhanced,drop shadow,colback=blue!5!]{$x_{d-1}$} & {\huge$\rightarrow$} & \tcbox[top=0pt,left=0pt,right=0pt,bottom=0pt,enhanced,sharp corners, drop shadow,colback=red!5!]{$h_{d-1}$} &  {\huge$\rightarrow$} &
  \tcbox[top=0pt,left=0pt,right=0pt,bottom=0pt,enhanced,drop shadow,colback=yellow!1!]{
    $p(x_d | x_{d-1}, \ldots)=p(x_d | h_{d-1})$ 
    \makecell{\includegraphics[width=0.1\textwidth]{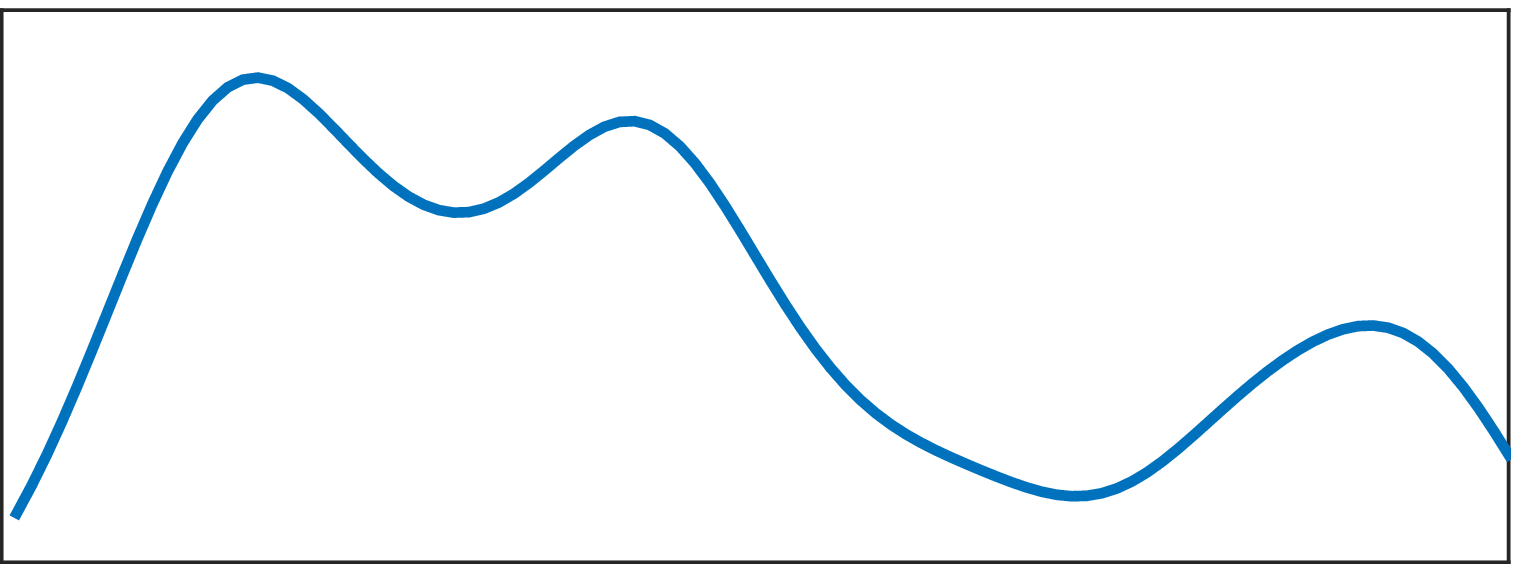}}
  }
  \\ 
  \vspace{0.1cm}\\ \hline
\end{tabular}
\end{center}
\vspace{0.1cm}
  \caption{Illustration of the use of recurrent neural networks for producing conditional distributions. Hidden states $h_k$'s are updated based previous values and then used to yield the parameters of the next dimension $x_{k+1}$. } \label{fig:conditionals}
\end{figure}

Consider now density estimation of a single multivariate variable $x\in \R^d$, a task in a similar vein to sequence estimation. General density estimation can be broken down into multiple conditional tasks on a growing set on inputs through the chain rule:
\begin{align}
p(x_1, \ldots x_d) = \prod_{i=1}^d p(x_i | x_{i-1}, \ldots, x_{1}).
\end{align}
As with sequential estimation, a completely general density estimation approach through the chain rule will suffer from the curse of dimensionality and a large number of seperate tasks. However, RNNs also provide an elegant solution in this case as one may model the density as:
\begin{align}
p(x_1, \ldots x_d) = \prod_{i=1}^d p(x_i | h_{i-1}) , \label{eq:RNN_cond}
\end{align}
where $h_i$ is the RNN state after observing the first $i$ dimensions, i.e. $h_i$ is a deterministic function of the previously seen dimensions $x_{i-1}, \ldots, x_{1}$ (see Figure~\ref{fig:conditionals}). In the case of gated-RNNs, the model will be able to scan through previously seen dimensions remembering and forgetting information as needed for conditional densities without making any strong Markovian assumptions.

Of course, unlike with sequential (or spatial) data, the indices of the dimensions of data may not have any semantic meaning. Thus, one is not tied to any intrinsic order in the use of the chain rule. Hence, a chain-rule approach may be framed as also depending on a permutation of one's data:
\begin{align}
p(x_1, \ldots x_d) = \prod_{i=1}^d p(x_{\pi_i} | x_{\pi_{i-1}}, \ldots, x_{\pi_1}). \label{eq:chain_perm}
\end{align}
In fact, due to the limited capacity of RNNs, a permutation $\pi$ that one considers may have a large impact on the estimation quality. One may be tempted to navigate the correlations among dimensions to find a suitable $\pi$, however the combinatorial nature of permutations makes such a task difficult and finding the optimal permutation will be infeasible.

Instead we note that just as one is not tied to any particular order, one is also not tied to a permutation transformation either.
Thus, rather than using a permutation $\pi$, we shall estimate conditionals of a transformation $z=(q_1(x), \ldots, q_d(x)) \in \R^d$:
\begin{align}
p(x_1, \ldots x_d) = \left|\det\frac{\ud z}{\ud x}\right| \prod_{i=1}^d p(q_i(x) | q_{i-1}(x), \ldots, q_1(x)),
\end{align}
where $\left|\det\frac{\ud z}{\ud x}\right|$ is a normalizing factor of the Jacobian of the transformation. 
By using RNNs for $q$ we shall get back a transformation that is flexible and easy to optimize; giving us generality with tractability.

\subsection{Conditional Densities}
As mentioned above we shall make use of a recurrent neural network to produce one-dimensional conditional distributions \eqref{eq:RNN_cond}. Recall that we make use of the RNN's hidden state after seeing the previous $i-1$ dimensions to produce the conditional density of the $i$th dimension (see Figure \ref{fig:conditionals}). We consider the use of gated recurrent units (GRUs) for this task because of their efficient parameterization and good performance relative to other recurrent units \cite{gru}; however, the use of other units is also possible. The update equations for the conditional density GRU are as follows:
\begin{gather}
    u_i = \sigma(x_{i}, h_{i-1}), \quad
    r_i = \sigma(x_{i}, h_{i-1}), \quad
    c_i = \sigma(x_{i}, r_i \odot h_{i-1}),\\
    h_i = u_i\odot h_{i-1} + (1-u_i)\odot h_{i-1},
    \label{eq:GRU}
\end{gather}
where we take $\sigma(\cdot, \cdot)$ to be an element-wise sigmoid applied to a linear projection of the arguments.
The hidden state $h_{i-1}$ \eqref{eq:GRU} determines the parameters of the 1d conditional density of the $i$th dimension, which we shall model as a mixture of Gaussians (GMM). For added nonlinearity we shall use two stacked fully connected units on $h_{i-1}$ to produce the means, $\theta^{\mu}_i$, the log of standard deviations $\theta^{\sigma}_i$, and the logits of mixture weights, $\theta^{w}_i$ for the GMM of the conditional density of the $i$th dimension.

\subsection{Transformations}

Simply iterating through the original dimensions of $x$ may lead to poor performance since there might not be any correlations present in adjacent dimensions and the conditional density RNN model may not be able to hold information from distantly seen dimensions. Hence, it may be useful to transform ones data before producing conditional densities. Ideally this transformation will be: one, easy to optimize; two, flexible enough to map points to a new space where dimensions are adept to being modeled in a recurrent manner. We accomplish such a transformation by stacking two types of invertable transformations together: a linear mapping, and recurrent transforms.

\begin{figure}[h]
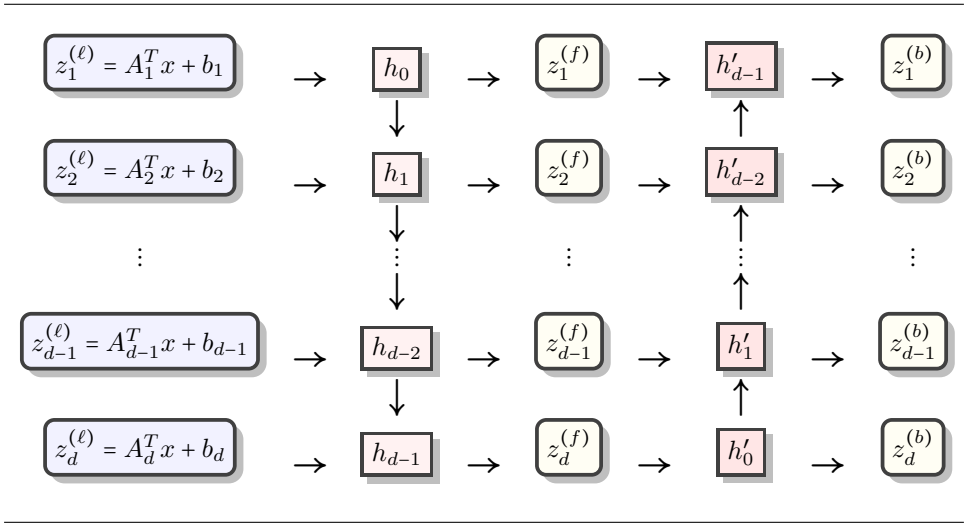

\begin{center}
\vspace{0.25cm}
\begin{tabular}{ c c c c c c c c c  } 
  \hline\\ 
   \tcbox[top=0pt,left=0pt,right=0pt,bottom=0pt,enhanced,drop shadow,colback=blue!5!]{$z^{(\ell)}_1 = A^T_1 x + b_1$} & {\LARGE$\rightarrow$} & \tcbox[top=0pt,left=0pt,right=0pt,bottom=0pt,enhanced,sharp corners,drop shadow,colback=red!5!]{$h_0$} &  {\LARGE$\rightarrow$} & \tcbox[top=0pt,left=0pt,right=0pt,bottom=0pt,enhanced,drop shadow,colback=yellow!5!]{$z^{(f)}_1$} &  {\LARGE$\rightarrow$} & \tcbox[top=0pt,left=0pt,right=0pt,bottom=0pt,enhanced,drop shadow,sharp corners,colback=red!10!]{$h^\prime_{d-1}$} &  {\LARGE$\rightarrow$} & \tcbox[top=0pt,left=0pt,right=0pt,bottom=0pt,enhanced,drop shadow,colback=yellow!5!]{$z^{(b)}_1$} \\
    &  & {\LARGE$\downarrow$} &  &  &  &  {\LARGE$\uparrow$}  &  \\
   \tcbox[top=0pt,left=0pt,right=0pt,bottom=0pt,enhanced,drop shadow,colback=blue!5!]{$z^{(\ell)}_2 = A^T_2 x + b_2$} & {\LARGE$\rightarrow$} & \tcbox[top=0pt,left=0pt,right=0pt,bottom=0pt,enhanced,drop shadow,sharp corners,colback=red!5!]{$h_1$} &  {\LARGE$\rightarrow$} & \tcbox[top=0pt,left=0pt,right=0pt,bottom=0pt,enhanced,drop shadow,colback=yellow!5!]{$z^{(f)}_2$} &  {\LARGE$\rightarrow$} & \tcbox[top=0pt,left=0pt,right=0pt,bottom=0pt,enhanced,drop shadow,sharp corners,colback=red!10!]{$h^\prime_{d-2}$} & {\LARGE$\rightarrow$} & \tcbox[top=0pt,left=0pt,right=0pt,bottom=0pt,enhanced,drop shadow,colback=yellow!5!]{$z^{(b)}_2$} \\
    &  & {\LARGE$\downarrow$} &  &  &  &  {\LARGE$\uparrow$}  &  \\
   $\vdots$ &  & $\vdots$ &  & $\vdots$ &  & $\vdots$ & & $\vdots$\\
      &  & {\LARGE$\downarrow$} &  &  &   &  {\LARGE$\uparrow$}  &  \\
   \tcbox[top=0pt,left=0pt,right=0pt,bottom=0pt,enhanced,drop shadow,colback=blue!5!]{$z^{(\ell)}_{d-1} = A^T_{d-1} x + b_{d-1}$} & {\LARGE$\rightarrow$} & \tcbox[top=0pt,left=0pt,right=0pt,bottom=0pt,enhanced,drop shadow,sharp corners,colback=red!5!]{$h_{d-2}$} &  {\LARGE$\rightarrow$} & \tcbox[top=0pt,left=0pt,right=0pt,bottom=0pt,enhanced,drop shadow,colback=yellow!5!]{$z^{(f)}_{d-1}$} & {\LARGE$\rightarrow$} & \tcbox[top=0pt,left=0pt,right=0pt,bottom=0pt,enhanced,drop shadow,sharp corners,colback=red!10!]{$h^\prime_{1}$} &  {\LARGE$\rightarrow$} & \tcbox[top=0pt,left=0pt,right=0pt,bottom=0pt,enhanced,drop shadow,colback=yellow!5!]{$z^{(b)}_{d-1}$} \\
    &  & {\LARGE$\downarrow$} &  &  &  &  {\LARGE$\uparrow$} &  \\
   \tcbox[top=0pt,left=0pt,right=0pt,bottom=0pt,enhanced,drop shadow,colback=blue!5!]{$z^{(\ell)}_d = A^T_d x + b_d$} & {\LARGE$\rightarrow$} & \tcbox[top=0pt,left=0pt,right=0pt,bottom=0pt,enhanced,drop shadow,sharp corners,colback=red!5!]{$h_{d-1}$} &  {\LARGE$\rightarrow$} & \tcbox[top=0pt,left=0pt,right=0pt,bottom=0pt,enhanced,drop shadow,colback=yellow!5!]{$z^{(f)}_d$} &  {\LARGE$\rightarrow$} & \tcbox[top=0pt,left=0pt,right=0pt,bottom=0pt,enhanced,drop shadow,sharp corners,colback=red!10!]{$h^\prime_{0}$} &  {\LARGE$\rightarrow$} & \tcbox[top=0pt,left=0pt,right=0pt,bottom=0pt,enhanced,drop shadow,colback=yellow!5!]{$z^{(b)}_{d}$} \\
  \vspace{0.1cm}\\ \hline
\end{tabular}
\end{center}
\vspace{0.1cm}
  \caption{Graphical representation of our transformation of variables for density estimation. First, we employ a linear mapping, $z^{(\ell)}=Ax+b$, to our input $x$ (where $A_k$ is the vector of the $k$th row of $A$). After, we use an invertible recurrent transformation with hidden states $h_k$ in a forward pass. Finally, we use another invertible recurrent transformation with hidden states $h^{\prime}_k$ in a backwards pass.  } \label{fig:transformation}
\end{figure}

First, we shall perform a linear transformation:
\begin{align}
z^{(\ell)} = A x + b,
\end{align}
where we take A to be invertable. Note that even though this linear transformation is simple, it includes permutations, and can thus shuffle the dimensions of $x$ for more favorable conditional estimation. Moreover, given the broader class of linear transformations over permutations, the mapping above may also perform a PCA-like transformation, capturing coarse and highly varied features of the data before moving to more fine grained details. In order to not incur a high cost for updates, we wish to compute the determinant of the Jacobian efficiently. We do so by directly working over an LU decomposition $A = LU$ where $L$ is a lower triangular matrix with unit diagonals and $U$ is a upper triangular matrix with arbitrary diagonals. As a function of $L$, $U$ we have that $\det\frac{\ud z^{(\ell)}}{\ud x} = \prod_{i=1}^d U_{ii}$; hence we may efficiently optimize the parameters of the linear map. Furthermore, inverting our mapping is also efficient through solving two triangular matrix equations.

Although a linear mapping is flexible enough to shuffle dimensions and capture different levels of variability, it is a transformation whose form does not depend on its input. That is, the dynamics of the linear model ($A$) remain constant regardless of $x$. While this leads to a simple model, there is a lack of flexibility, since one may wish to transform the input differently depending on its values.
However, it is important to not sacrifice efficiency for the sake of flexibility, thus we still desire to have a fast transformation in terms of computation, inversion, and learning.  

Recurrent neural networks are also a natural choice for variable transformations. Due to their dependence on only previously seen dimensions, RNN transformations have triangular Jacobian, leading to simple determinants. Furthermore, with an invertible output unit, their inversion is also straight-forward. We consider the following form to RED RNN transformations (for ease of notation take the input to the recurrent transformation to be $x$):
\begin{align}
    z_i = r_\alpha\left( y x_i + w^{T} h_{i-1} + b \right),\quad
    h_i = r\left( u x_i + v^{T} h_{t-1} + a \right), \label{eq:rnn_trans}
\end{align}
where $r_\alpha$ is a leaky ReLU unit $r_\alpha(s) = \I\{s<0\} \alpha s + \I\{s\ge0\} s$, $r$ is a standard ReLU unit,  $y$, $u$, $b$ $a$ are scalars, and $w$, $v$ as vectors.
Inverting \eqref{eq:rnn_trans} is a matter of inverting outputs and updating the hidden state (where the initial state $h_0$ is known and constant):
\begin{align}
    x_i = y^{-1}\left(r^{-1}_\alpha\left(z^{(r)}_i\right) - w^{T} h_{i-1} - b\right),\quad
    h_i = r\left( u x_i + v^{T} h_{t-1} + a \right). \label{eq:rnn_trans_inv}
\end{align}
Furthermore, the determinant of the Jacobian for \eqref{eq:rnn_trans} is the product of diagonal terms:
\begin{align}
    \det\frac{\ud z}{\ud x} = y^d \prod_{i=1}^d  r^{\prime}_\alpha\left( y x_i + w^{T} h_{i-1} + b \right),
\end{align}
where $r^{\prime}_\alpha\left(t\right) = \I\{t> 0\} + \alpha \I\{t<0\}$. 
For added dependence among all dimensions, we consider a forward pass of a recurrent transformation \eqref{eq:rnn_trans} followed by another recurrent transformation in a backwards pass. 

Notwithstanding their natural fit, this is, to the best of our knowledge, the first use of recurrent networks for transformations of variables in general data density estimation.
In total we consider three transformations to covariates, an initial linear mapping, then forwards and backwards passes of recurrent transformations: $x \mapsto z^{(\ell)} \mapsto z^{(f)} \mapsto z^{(b)}$
(see Figure~\ref{fig:transformation}).

Lastly, we note that sampling a RED model is computationally efficient. One must simply propagate random draws of dimensions through the conditional RNN: $\ddot{z}^{(b)}_{k} \sim p({z}^{(b)} | \ddot{z}^{(b)}_{k-1}, \ldots, \ddot{z}^{(b)}_{1}) = p({z}^{(b)} | \ddot{h}_{k-1})$, where $\ddot{z}^{(b)}_{j}$ are the drawn dimensions of the latent space and $\ddot{h}_{k-1}$ is the corresponding conditional hidden state after seeing $\ddot{z}^{(b)}_{k-1}, \ldots, \ddot{z}^{(b)}_{1}$. After, one inverts the linear/RNN transformation to produce a sample, $\ddot{x} = q^{-1}(\ddot{z}^{(b)})$. As explained above, inverting this transformation is inexpensive, especially when caching $A^{-1}$ (a one time cost of inverting the product of two triangular matrices $L, U$).

\section{Related Work}
Nonparametric density estimation has been a well studied problem in statistics and machine learning \cite{larry}. Unfortunately, nonparametric approaches like kernel density estimation suffer greatly from the curse of dimensionality and do not perform well when data does not have a small number of dimensions ($\lesssim 3d$). To alleviate this, several semiparametric approaches have been explored. Such approaches include forest density estimation \cite{lafferty}, which assumes that the data has a forest (i.e. a collection of trees) structured graph. This assumption leads to a density which factorizes in a first order Markovian fashion through a tree traversal of the graph. Another common semiparametric approach is to use a nonparanormal type model \cite{liu}. This approach uses a Gaussian copula with a rank-based transformation and a sparse precision matrix. While both approaches are well-understood theoretically, their strong assumptions often lead to inflexible models.

In order to provide greater flexibility with semiparametric models, recent work has employed deep learning for density estimation. The use of neural networks for density estimation dates back to early work by \cite{bishop} and has seen success in areas like speech \cite{speech, Uriaspeech}, music \cite{NicolasMusic}, etc.. Typically such approaches use a network to learn to parameters of a parametric model for data. Recent work has also explored the application of deep learning to build density estimates in image data \cite{pixel, dinh2}. However, such approaches are heavily reliant on exploiting structure in neighboring pixels, often reshaping or re-ordering data and using convolutions to take advantage of neighboring correlations. Modern approaches for general density estimation in real-valued data include neural autoregressive density estimation (NADE) \cite{uria1, uria2} and non-linear independent component estimation (NICE) \cite{dinh1}. We briefly describe these approaches below and contrast them to RED models.

\subsection{NADE}
NADE is a restricted-Boltzmann-machine-inspired density estimator that uses probability product rules and a weight-sharing scheme across conditional densities directly on input covariates \cite{uria2}. 
As previously mentioned, a direct application of the chain rule on input dimensions will be sensitive to the order used \eqref{eq:chain_perm}.
To overcome the difficulty of finding a good permutation, deep versions of NADE train the model by sampling random permutations, $\pi$, throughout training. I.e. NADE attempts to find parameters that model the data well (on average) for all permutations. Of course, due to the combinatorial nature of permutations, only a minuscule fraction of permutations will have been explored during training time.

Given that NADE is devoid of a latent space transformation of variables, it may not find a space that is adept to conditional modeling; instead it is limited to only working in the original covariates, which may contain complex dependencies that might not be captured with a simple hidden state representation. 
Furthermore, in comparison to NADE, RED models have the ability to update hidden states recursively based on previous values.


\subsection{NICE}
NICE models assume that data is drawn from a latent independent Gaussian space and transformed \cite{dinh1}. The transformation uses several ``additive coupling'' shifting transformations on the second half of dimensions, using the first half of dimensions. The full transformation is the result of stacking several of these additive coupling layers together followed by a final rescaling operation. The resulting composition of these transformation yields a simple determinant of the Jacobian. 

By assuming independence in the latent space, NICE models must entirely rely on the deterministic transformation to capture correlations.
In contrast, RED models consider conditional structure in a latent space directly and further refine drawn data with a deterministic transformation afterwards.

\section{Experiments}
We compare NICE, NADE, and RED models using several experiments on real-world datasets. First, we compute the average negative-log-likelihoods across instances of held out test data. Then, to gain further context of the efficacy of models, we also use their density estimates for anomaly detection, where we take low density instances to be outliers. 

\subsection{Methodology}
We implemented RED using Tensorflow \cite{tensorflow}, making use of the standard \texttt{GRUCell} GRU implementation\footnote{Code will be made public upon publication.}. We used author-provided code for NADE \cite{nade_code} and NICE \cite{nice_code} models. In order to ensure a fair exploration over hyper-parameters, a grid search was performed over several choices for the following parameters for models:
\texttt{num\_units}, size of hidden layers in models;
\texttt{init\_lr}, the initial learning rate to use for optimization;
\texttt{decay\_factor}, parameter controlling how to decay the learning rate;
\texttt{num\_fcs} number of fully connected layers per coupling transformation in NICE, or the number of layers applied to hidden states for conditional densities in RED and NADE; 
\texttt{num\_components}\footnote{Only used in RED and NADE models}, number mixture components for conditional densities; 
\texttt{min\_lr\_factor}\footnote{\label{fnote:nice}Only used for NICE models.}\addtocounter{footnote}{-1}\addtocounter{Hfootnote}{-1}, a multiplicative factor of \texttt{init\_lr} that determines minimum learning rate to consider;
\texttt{num\_coupling}\footnotemark, number of coupling layers to use.


\subsection{Datasets}
We used multiple datasets from the UCI machine learning repository \cite{uci}, Stony Brook outlier detection datasets collection (ODDS) \cite{odds}, as well as other sources to evaluate log-likelihoods on held-out test data. Broadly, the datasets can be divided into:
\begin{itemize}
    \item \textbf{Particle acceleration}: \texttt{higgs}, \texttt{hepmass}, and \texttt{susy} datasets where generated for high-energy physics experiments using Monte Carlo simulations. 
    \item \textbf{Song}: The \texttt{song} dataset contains timbre features from the million song dataset of mostly commercial western song tracks from the year 1922 to 2011 \cite{millisongs}.
    \item \textbf{Word2Vec}: \texttt{wordvecs} consists of 3 million words from a Google News corpus. Each word represented as a 300 dimensional vector trained using a word2vec model \cite{word2vec}.
    \item \textbf{Outlier detection datasets}: We also used several ODDS datasets--\texttt{shuttle}, \texttt{forest}, \texttt{pendigits}, \texttt{satimage2}, \texttt{speech}. These are multivariate datasets from varied set of sources meant to provide a complete picture of performance across anomaly detection tasks. We note that in order to not penalize models for low likelihoods on outliers, we removed anomalies from the test set when reporting log-likelihoods.
\end{itemize}
As noted in \cite{dinh1}, data degeneracies and other corner-cases may lead to arbitrarily low negative log-likelihoods. In order to avoid such complications, we first standardized all datasets, and added independent Gaussian noise with a standard deviation of $0.01$ to training sets. 

\subsection{Test-Data Log Likelihoods}
We report average negative log-likelihoods on test data in Table~\ref{tbl:nlls}.
RED models consistently provide a lower negative log likelihood on test data,
and outperform both NADE and NICE on almost all datasets. Furthermore, it is interesting to note that this trend holds true across numerous different dataset sizes and dimensionalities. This suggests that RED can adapt to a myriad of circumstances. Moreover, there does not seem to be a clear winner between NICE and NADE, suggesting that perhaps each respective model is lacking in different situations, whereas RED seems to adjust well throughout.

\begin {table}[h]
  \begin{center}
  \begin{tabular} { |c|c|c|c|c|c| }
    \hline
    Dataset & N & d & RED NLL & NICE NLL & NADE NLL \\ \hline
    \texttt{shuttle} & 49,097 & 9 & \textbf{-10.30} & 50.14 &	-7.74 \\
    \texttt{forest} & 286,048 & 10 & \textbf{0.20} & 1.23 & 1.33 \\
    \texttt{pendigits} & 6,870 & 16 & -3.29 & 5.65 &	\textbf{-5.05} \\
    \texttt{susy} & 5,000,000 & 18 & \textbf{-9.72} &	-2.87 &	4.08 \\
    \texttt{higgs} & 11,000,000 & 28 & \textbf{12.18} & 18.87 & 12.28 \\ 
    \texttt{hepmass} & 10,500,000 & 28 & \textbf{3.35}	& 15.93 & 4.21 \\    
    \texttt{satimage2} & 5,803 & 36 & \textbf{0.64} & 3.14 & 2.45 \\
    \texttt{song} & 515,345 & 90 & \textbf{74.92} &	79.37 &	75.31 \\
    \texttt{wordvecs} & 3,000,000 & 300 & \textbf{294.26} & 363.03 & 296.61 \\
    \texttt{speech}	& 3,686 & 400 & \textbf{555.54} & 561.14 & 561.85 \\
    \hline
  \end{tabular}
  \end{center}
  \caption{Average negative log-likelihood on held-out test data. Lowest values per dataset are shown in \textbf{bold}. All reported lowest values are statistically significantly lower ($\rho < 0.001$) than the second best with 
  a paired-T test. We also report the number of instances and dimensions per dataset, $N$ and $d$, respectively. }
  \label{tbl:nlls}
\end{table}


\subsection{Anomaly Detection}
Next, we apply density estimates to anomaly detection. Typically anomalies or outliers are data-points that are unlikely given a dataset. In terms of density estimations, such a task is framed by identifying which instances in a dataset have a low corresponding density. That is, we shall label an instance $x$, as an anomaly if $\hat{p}(x) \leq t$, where $t\geq 0$ is some threshold and $\hat{p}$ is the density estimate based on training data. Note that this approach is trained in an unsupervised fashion. However, each methods' density estimates were evaluated on test data with anomaly/non-anomaly labels on instances. We used thresholded log-likelihoods on the test set to compute precision and recall. We use the average-precision metric:
\begin{align}
    \text{avg-prec} = \sum_{k=1}^{N_{\mathrm{test}}}\text{precision}_r \left (\text{recall}_r - \text{recall}_{r-1} \right ) 
     \label{eq:avgprec}
\end{align}
where $\text{precision}_r = \frac{tp_r}{tp_r+fp_r}$, $\text{recall}_r = \frac{tp_r}{tp_r+fn_r}$, and $tp_r, fp_r, fn_r$ are true positive anomalies, false positives and false negative respectively among the bottom $r$ log-likelihood instances in test data.

\begin {table}[h]
  \begin{center}
  \begin{tabular} { |c|c|c|c|c| }
    \hline
    Dataset & Anomaly Count & RED avg-prec & NICE avg-prec  & NADE avg-prec \\ \hline
    \texttt{shuttle} & 3,511 & \textbf{0.999} & 0.996 & 0.997 \\
    \texttt{forest} & 2,747 & \textbf{0.803}	& 0.781 & 0.536 \\
    \texttt{pendigits} & 156 & \textbf{0.993} &	0.933 &	0.973 \\
    \texttt{satimage2} & 71  & \textbf{0.992} &	0.983 &	0.991 \\    
    \texttt{speech}	&  61 &	\textbf{0.193} &	0.190 &	0.156 \\
     \hline
     \multicolumn{2}{|c|}{ \emph{MAP}} & \textbf{0.796} & .777 & .731 \\
     \hline
     \multicolumn{2}{|c|}{ \emph{nDCG }} & \textbf{0.790} & .782 & .73 \\
     \hline 
  \end{tabular}
  \end{center}
  \caption{Average precision  on anomaly detection tasks \label{tab:outlier}}
\end{table}

\begin{figure}[h]
    \centering
    \begin{subfigure}[b]{0.3\textwidth}
        \includegraphics[width=\textwidth]{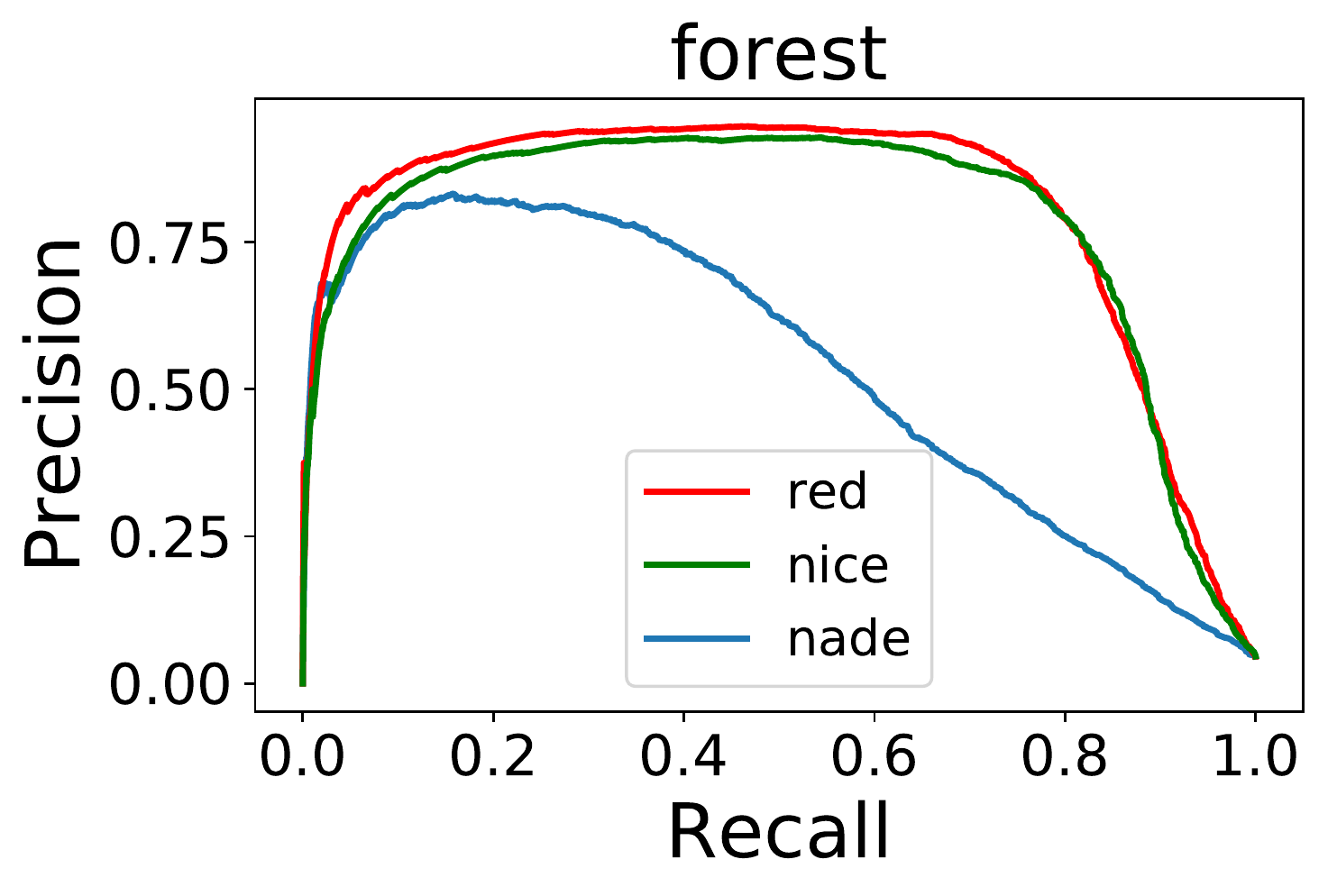}
    \end{subfigure}
    \begin{subfigure}[b]{0.3\textwidth}
        \includegraphics[width=\textwidth]{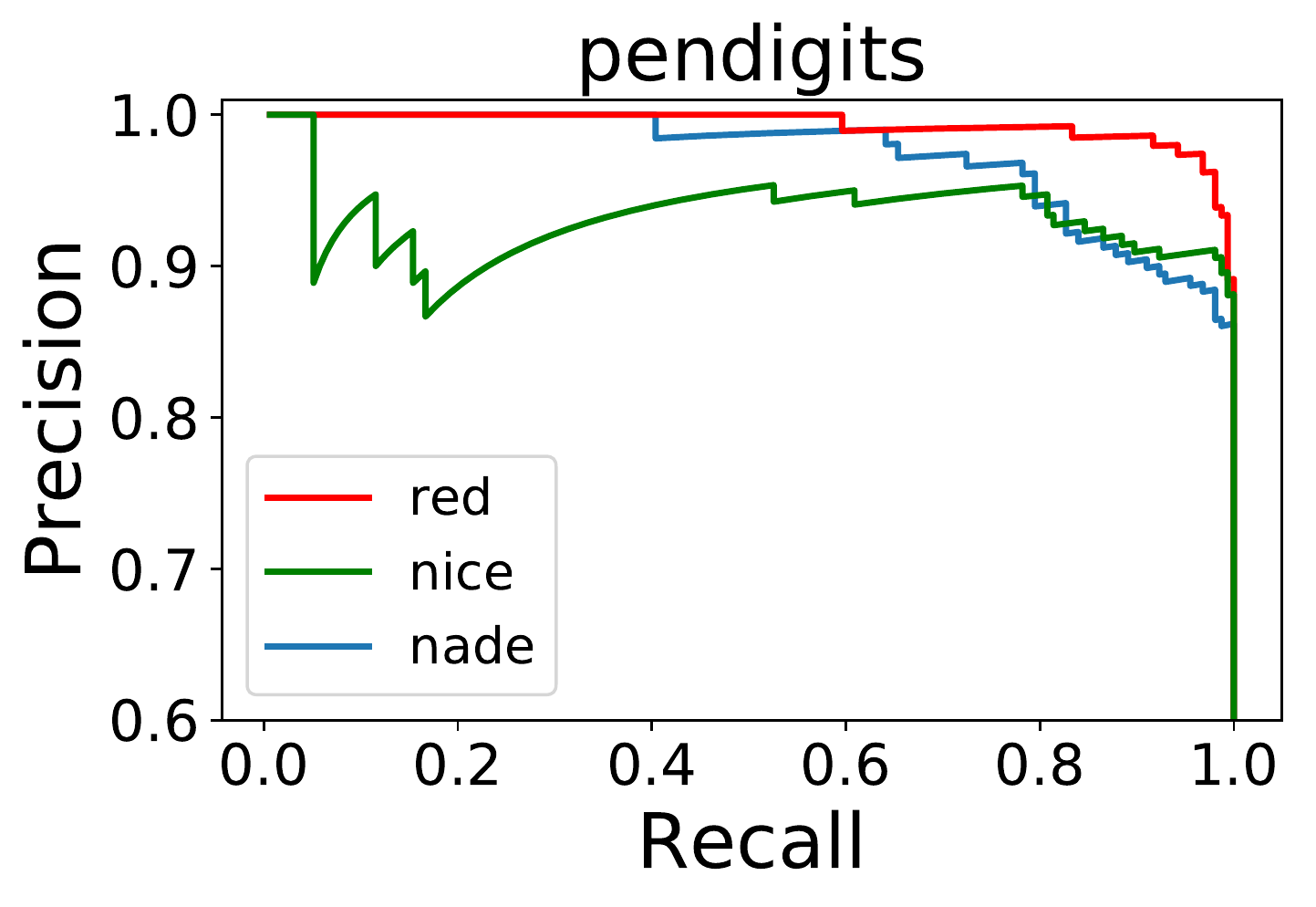}
    \end{subfigure}
    \begin{subfigure}[b]{0.3\textwidth}
        \includegraphics[width=\textwidth]{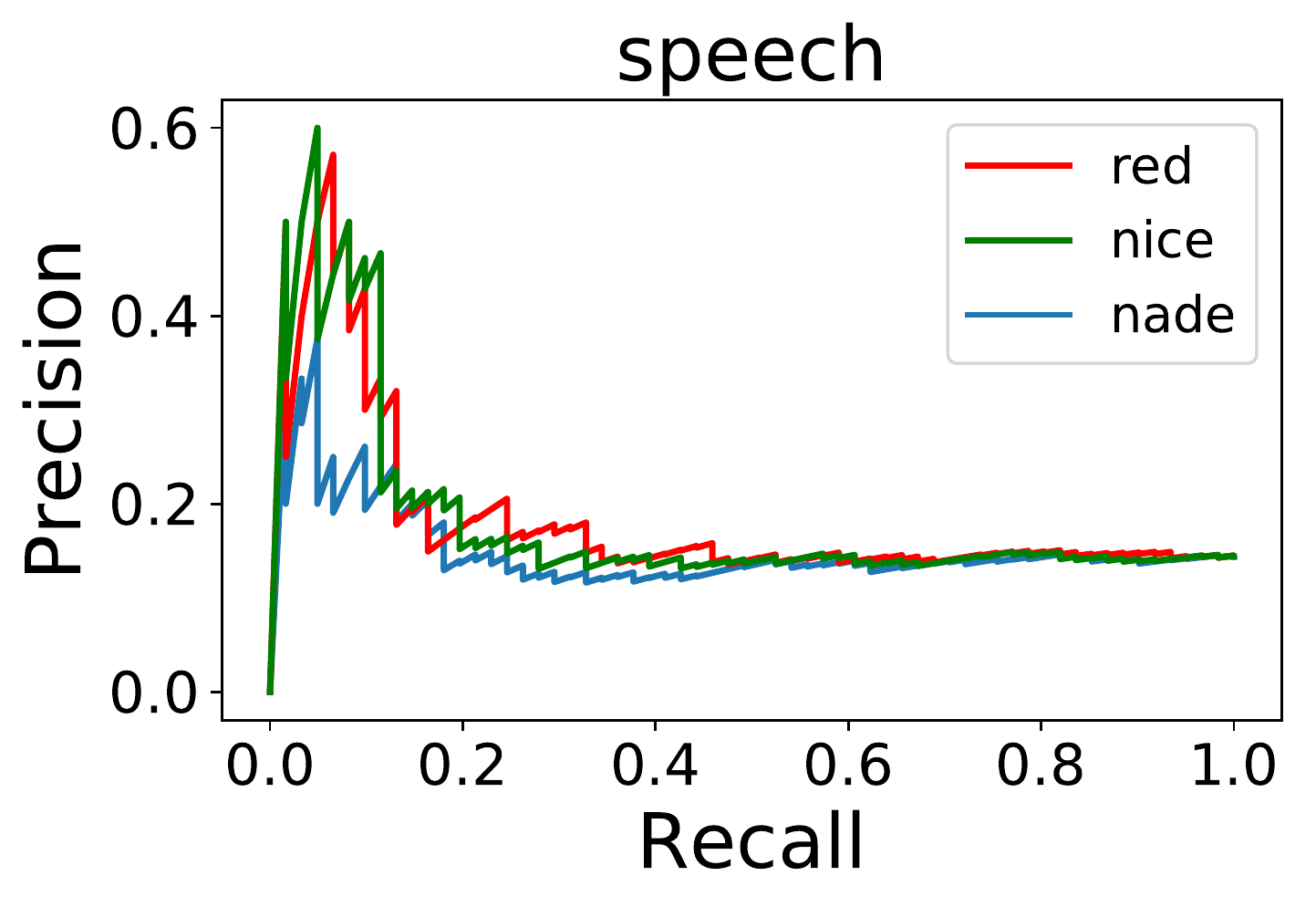}
        
    \end{subfigure}
    \caption{Precision-Recall curve for outlier detection. Precision and recall values have been calculated by thresholding at the bottom $r$ ranked log-likelihood instances in test data and varying $r$ from $1$ to $N_{test}$. }
    \label{fig:prec-recall}
\end{figure}

Our results are shown in Table~\ref{tab:outlier}. We see that RED also outperforms NICE and NADE methods for outlier detection. In order to provide a holistic measure of performance across all dataset, we also report two measures: \emph{mean average precision} (MAP) and \emph{normalized discounted cumulative gain} (nDCG) (see Ch-8 of \cite{Manning} for details).
To visualize the performance on specific datasets, we plot the precision-recall curves in Figure~\ref{fig:prec-recall}. One can see that RED tends to achieve a higher precision across recall values.

\vspace{-0.2cm}

\section{Discussion}
In conclusion, this work presents the recurrent estimation of distributions, a novel two-prong usage of recurrent neural networks for density estimation with general real-valued data. First, RNNs are used to transform input covariates into a latent space more adept to capturing dependencies. This data-driven approach to ``pre-pocessing'' covariates is used to exploit correlations in data without hard coding any dependencies to dimensions as is common with spatial/temporal-specific models. After, we apply an RNN to model conditional densities across the transformed dimensions. Gated-RNNs are especially adept for this task as they may scan through the data remembering and forgetting information as needed without having to make a strong Markovian assumptions. 

The efficacy of the RED method as compared to other neural network approaches was shown on various datasets. We find that RED models achieve a lower negative log-likelihood across multiple dataset sizes and dimensionalities. 
Furthermore, the fact that there is no clear winner between alternatives NADE and NICE suggest that capturing both latent space conditional dependencies and change of variable transformations are key for achieving top performance.
Lastly, we apply density estimates to anomaly detection and also find that RED provides superior performance for multiple datasets.

\clearpage

\bibliography{main}

\begin{thebibliography}{10}

\bibitem{nade_code}
{NADE code}.
\newblock \url{https://github.com/MarcCote/NADE}.
\newblock Accessed: 04-01-2017.

\bibitem{nice_code}
{NICE code}.
\newblock \url{https://github.com/laurent-dinh/nice}.
\newblock Accessed: 04-01-2017.

\bibitem{odds}
{Outlier Detection DataSets (ODDS)}.
\newblock \url{http://odds.cs.stonybrook.edu}.
\newblock Accessed: 04-01-2017.

\bibitem{uci}
{UCI Machine Learning Repository}.
\newblock \url{http://archive.ics.uci.edu/ml/}.
\newblock Accessed: 04-01-2017.

\bibitem{word2vec}
{word2vec}.
\newblock \url{https://code.google.com/archive/p/word2vec/}.
\newblock Accessed: 04-01-2017.

\bibitem{tensorflow}
Mart{\'\i}n Abadi, Ashish Agarwal, Paul Barham, Eugene Brevdo, Zhifeng Chen,
  Craig Citro, Greg~S Corrado, Andy Davis, Jeffrey Dean, Matthieu Devin, et~al.
\newblock Tensorflow: Large-scale machine learning on heterogeneous distributed
  systems.
\newblock {\em arXiv preprint arXiv:1603.04467}, 2016.

\bibitem{millisongs}
Thierry Bertin-Mahieux, Daniel~PW Ellis, Brian Whitman, and Paul Lamere.
\newblock The million song dataset.
\newblock In {\em ISMIR}, volume~2, page~10, 2011.

\bibitem{bishop}
Christopher~M Bishop.
\newblock Mixture density networks.
\newblock {\em Technical Report}, 1994.

\bibitem{NicolasMusic}
N.~Boulanger-Lewandowski, Y.~Bengio, and P.~Vincent.
\newblock Modeling temporal dependencies in high-dimensional sequences:
  Application to polyphonic music generation and transcription.
\newblock {\em International Conference on Machine Learning}, 2012.

\bibitem{gru}
Junyoung Chung, Caglar Gulcehre, KyungHyun Cho, and Yoshua Bengio.
\newblock Empirical evaluation of gated recurrent neural networks on sequence
  modeling.
\newblock {\em arXiv preprint arXiv:1412.3555}, 2014.

\bibitem{dinh1}
Laurent Dinh, David Krueger, and Yoshua Bengio.
\newblock Nice: Non-linear independent components estimation.
\newblock {\em arXiv preprint arXiv:1410.8516}, 2014.

\bibitem{dinh2}
Laurent Dinh, Jascha Sohl-Dickstein, and Samy Bengio.
\newblock Density estimation using real nvp.
\newblock {\em arXiv preprint arXiv:1605.08803}, 2016.

\bibitem{liu}
Han Liu, John Lafferty, and Larry Wasserman.
\newblock The nonparanormal: Semiparametric estimation of high dimensional
  undirected graphs.
\newblock {\em Journal of Machine Learning Research}, 10(Oct):2295--2328, 2009.

\bibitem{lafferty}
Han Liu, Min Xu, Haijie Gu, Anupam Gupta, John Lafferty, and Larry Wasserman.
\newblock Forest density estimation.
\newblock {\em Journal of Machine Learning Research}, 12(Mar):907--951, 2011.

\bibitem{Manning}
C.~D. Manning, Raghavan P., and H.~Schütze.
\newblock Introduction to information retrieval, 2009.

\bibitem{pixel}
Aaron van~den Oord, Nal Kalchbrenner, and Koray Kavukcuoglu.
\newblock Pixel recurrent neural networks.
\newblock {\em arXiv preprint arXiv:1601.06759}, 2016.

\bibitem{Uriaspeech}
B.~Uria.
\newblock Connectionist multivariate density-estimation and its application to
  speech synthesis., 2015.

\bibitem{uria2}
Benigno Uria, Marc-Alexandre C{\^o}t{\'e}, Karol Gregor, Iain Murray, and Hugo
  Larochelle.
\newblock Neural autoregressive distribution estimation.
\newblock {\em Journal of Machine Learning Research}, 17(205):1--37, 2016.

\bibitem{uria1}
Benigno Uria, Iain Murray, and Hugo Larochelle.
\newblock A deep and tractable density estimator.
\newblock In {\em ICML}, pages 467--475, 2014.

\bibitem{larry}
L~Wasserman.
\newblock All of nonparametric statistics, 2007.

\bibitem{speech}
Heiga Zen and Andrew Senior.
\newblock Deep mixture density networks for acoustic modeling in statistical
  parametric speech synthesis.
\newblock In {\em Acoustics, Speech and Signal Processing (ICASSP), 2014 IEEE
  International Conference on}, pages 3844--3848. IEEE, 2014.

\end{thebibliography}
\bibliographystyle{plain}

\end{document}